\documentclass[letterpaper, 10 pt, conference]{ieeeconf}  

\IEEEoverridecommandlockouts                              

\usepackage{amsmath} 
\usepackage{amssymb}  
\usepackage{biblatex} 
\usepackage{booktabs}
\usepackage{multirow}
\usepackage{graphicx}
\usepackage{subfigure}
\DeclareMathOperator*{\argmax}{argmax}

\title{\LARGE \bf
RGB Matters: Learning 7-DoF Grasp Poses on \\Monocular RGBD Images
}

\author{Minghao Gou$^{1}$, Hao-Shu Fang$^{2}$, Zhanda Zhu$^{2}$, Sheng Xu$^{1}$, Chenxi Wang$^{1}$ and Cewu Lu$^{1}$\\
Shanghai Jiao Tong University
\thanks{$^{1}$\{gmh2015, xs1020, wcx1997, lucewu\}@sjtu.edu.cn. Cewu Lu is the corresponding author, member of Qing Yuan Research
Institute and MoE Key Lab of Artificial Intelligence, AI Institute, Shanghai
Jiao Tong University, China}
\thanks{$^{2}$ \{fhaoshu, zhandazhu\}@gmail.com}%
}

\begin{document}

\maketitle
\thispagestyle{empty}
\pagestyle{empty}

\begin{abstract}
General object grasping is an important yet unsolved problem in the field of robotics. Most of the current methods either generate grasp poses with few DoF that fail to cover most of the success grasps, or only take the unstable depth image or point cloud as input which may lead to poor results in some cases. In this paper, we propose RGBD-Grasp, a pipeline that solves this problem by decoupling 7-DoF grasp detection into two sub-tasks where  RGB and depth information are processed separately. In the first stage, an encoder-decoder like convolutional neural network \emph{Angle-View Net}(AVN) is proposed to predict the SO(3) orientation of the gripper at every location of the image. Consequently, a \emph{Fast Analytic Searching}(FAS) module calculates the opening width and the distance of the gripper to the grasp point. By decoupling the grasp detection problem and introducing the stable RGB modality, our pipeline alleviates the requirement for the high-quality depth image and is robust to depth sensor noise. We achieve state-of-the-art results on GraspNet-1Billion dataset compared with several baselines. Real robot experiments on a UR5 robot with an Intel Realsense camera and a Robotiq two-finger gripper show high success rates for both single object scenes and cluttered scenes. Our code and trained model will be made publicly available.

\end{abstract}

\section{INTRODUCTION}\label{sec:introduction}
\vspace{-0.05in}
Grasping is a fundamental and important problem in robotics which is the basis for robotic manipulation. Despite its vital importance, solutions to this problem are far from satisfactory. 

Traditional methods~\cite{traditional1}~\cite{traditional2}~\cite{traditional3} utilize physical analysis to find suitable grasp poses. However, these methods require accurate object models which are not always available. It's also difficult to apply these algorithms to unseen objects. Besides, these methods are usually time consuming and computationally expensive.

With the development of computer vision and artificial intelligence, researchers propose data-driven or learning based methods to solve this problem.
2D planar grasping methods are firstly studied. Several datasets~\cite{cornell}~\cite{supersizing}~\cite{jacquard} are proposed and many algorithms ~\cite{cornell}~\cite{lenz}~\cite{ggcnn}~\cite{2014redmon}~\cite{kumra}~\cite{asif}~\cite{dsgd}~\cite{multi} learn to generate 2D planar grasp poses on those datasets. A few of them get a high accuracy on 2D grasping metrics. However, 2D planar grasping model imposes many restrictions on the grasp poses. The gripper can only approach the object in the top-down direction while in some cases it fails to grasp the object along this direction. For example, it's very difficult for a gripper to grab a horizontally placed plate.

Recently, researchers begin to explore 6-DoF grasp poses detection. 6D pose estimation~\cite{peng2019pvnet}~\cite{he2020pvn3d}~\cite{zhao2018estimating} allows 6-DoF grasping on known objects but cannot generalize to novel scene. GPD~\cite{gpd} and PointnetGPD~\cite{pointnetgpd} follow a two-step sampling-evaluation method. However, due to the unsatisfactory quality of the sampled results, a huge number of samples have to be evaluated to find reliable grasp poses which is time-consuming.  Tian \emph{et al.}~\cite{tian2019transferring}, Florence \emph{et al.}~\cite{denseobjectnets} and Patten \emph{et al.}~\cite{dgcmnet} transfer grasp poses from existing ones. But these methods fail when the objects are novel and the geometries are not similar to any existing one. Fang \emph{et al.}~\cite{graspnet}, Ni \emph{et al.}~\cite{pointnetppgrasping} and Mousavian \emph{et al.}~\cite{6dofgraspnet} feed partial-view point cloud captured by RGBD camera to neural networks to obtain 6-DoF grasp poses. However, depth data is not stable compared with RGB data because of potential sensor failure. 
\begin{figure}[t]
	\centering
	\includegraphics[width=0.50\textwidth]{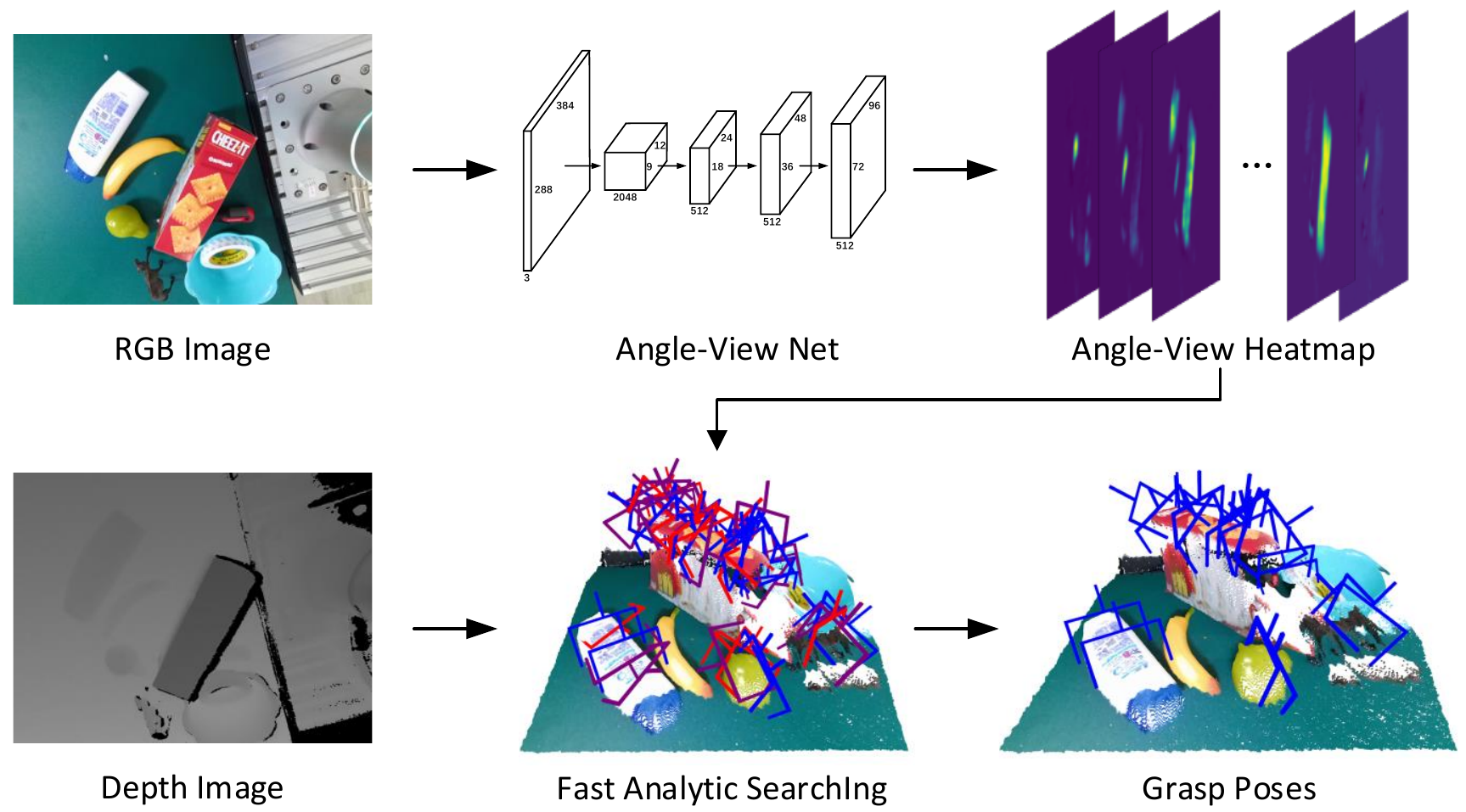}
	\caption{An overview of our method. Top Left: The input RGB image. Top Middle: The \emph{Angle-View Net}(\textbf{AVN}). Top Right: Output \emph{Angle-View Heatmap}s(\textbf{AVH}s), Bottom Left: The input depth image. Bottom Middle: The collsion and empty grasp detection based \emph{Fast Analytic Searching}(\textbf{FAS}) module. Bottom Right: The output 7-DoF grasp poses.}
	\label{fig:introduction}
\end{figure}

In this paper, we present RGBD-Grasp, a 7-DoF grasp detection pipeline. We take a monocular RGBD image as input and output the 6-DoF gripper pose alongside with the width of gripper as an additional DoF. As shown in Figure~\ref{fig:introduction}, we decouple the problem into two sub-problems. Using only the RGB image, we generate a group of heatmaps that indicate the pixel-wise orientation matrix of the gripper by an encoder-decoder network. Then we calculate the gripper opening width and the distance to the grasp point by collision and empty grasp detection guided fast analytic searching using the heatmap and the depth image.

The advantages of RGBD-Grasp are threefold. Firstly, it is the first attempt to generate high DoF grasp poses using a monocular RGBD image. The use of RGB image makes the method more stable and high DoF grasp poses solve the problem of planar restriction. Secondly, although the method is composed of two stages, each of the stages runs efficiently. In our experiment, it runs at 10.5 FPS which meets the requirement of real-time operation. Thirdly, benefiting from the deep CNN and large-scale data of GraspNet-1Billion dataset~\cite{graspnet}, the method not only works well with seen or similar objects, but it also generalizes to novel objects.

We evaluate our method on the GraspNet-1Billion benchmark~\cite{graspnet}. Compared with existing methods, we improve the AP by \textbf{2.20}, \textbf{2.56} and \textbf{1.57} for seen, unseen and novel objects respectively for Kinect and \textbf{0.42}, \textbf{1.12} and \textbf{1.70} for RealSense and achieve the state-of-the-art. We also conduct real robot experiments for both single object scenes and cluttered scenes with a UR5 robot mounted with a RealSense camera. We report \textbf{91.7\%} of the average success rate for single object scenes and \textbf{91.1\%} for cluttered scenes.
\vspace{-0.02in}
\section{RELATED WORKS}\label{sec:relatedworks}
\vspace{-0.02in}
In this section, we focus on reviewing deep learning based grasping methods. Several review papers~\cite{review1}~\cite{review2}~\cite{review3}~\cite{review4} have made detailed analysis on the works in this field. According to the representation of grasp poses, previous works can be divided into mainly two types: 2D planar grasping and 6-DoF grasping.
\vspace{-0.05in}
\subsection{2D Planar Grasping}\label{subsec:2dgrasping}
2D planar grasping means the gripper is constrained in the direction perpendicular to the camera plane. In this case, the grasp pose is often represented by an oriented rectangle which gives the location, orientation, width and height of the gripper from the top-down view.
Jiang \emph{et al.}~\cite{cornell} propose the Cornell dataset of 2D planar grasping and a searching based algorithm to find the best grasp pose. However, both the speed and performance are far from satisfactory.
Lenz \emph{et al.}~\cite{lenz} search over a large space to generate several candidates and use a CNN to process both RGB and depth images to evaluate them. The top ranked grasp is selected for execution.
Mahler \emph{et al.}~\cite{dexnet2} generate many 2D grasp candidates and evaluate them with  GQ-CNN.
Morrison \emph{et al.}~\cite{ggcnn} propose GG-CNN to predict the quality and pose of grasps at each pixel.
Redmon \emph{et al.}~\cite{2014redmon} and Kumra \emph{et al.}~\cite{kumra} directly regress the parameters for 2D planar grasp and grid the picture to obtain multiple grasp poses.
Asif \emph{et al.} ~\cite{asif} propose an encoder-decoder network to generate an affordance map and calculate the 2D grasp poses accordingly.
Inspired by Region Proposal Network~\cite{fasterrcnn}, Chu \emph{et al.} ~\cite{multi} propose Grasp Proposal Network to detect a lot of grasps on multi-object images.
Asif \emph{et al.} ~\cite{dsgd} assemble three different types of networks and switch among them according to their confidence scores. 
Although the reported scores in these papers are high, the constraint of 2D planar grasp is strong so that many suitable grasps cannot be represented in this format, which limits their application.

\subsection{6-DoF Grasping}\label{subsec:6dgrasping}
For 6-DoF grasping methods, the gripper can grasp objects from arbitrary directions. Six parameters are needed to specify the 3D location and rotation. Additional DoFs such as the width or height of the gripper may also be part of the representation. 
GPD~\cite{gpd} and PointnetGPD~\cite{pointnetgpd} sample grasps in the space with constraint and then use LeNet or Pointnet to evaluate these samples respectively. 
Mousavian \emph{et al.}~\cite{6dofgraspnet} sample grasp poses using a variational autoencoder and then assess and refine the sampled poses.
S$^4$G~\cite{s4g} takes a partial view point cloud as input and regresses the 6-DoF poses directly.
Florence \emph{et al.}~\cite{denseobjectnets}, DGCM-Net~\cite{dgcmnet} and Tian \emph{et al.}~\cite{tian2019transferring} transfer existing grasp poses from one object to another.
Yan \emph{et al.}~\cite{yan2018learning}, Lundell \emph{et al.}~\cite{Lundell} and Varley \emph{et al.}~\cite{shapecompletion} firstly reconstruct the scene from partial view point cloud and then plan grasps on the completed scene.
Recently, Fang \emph{et al.}~\cite{graspnet} propose the first large-scale dataset in this area and decouple the grasp poses estimation into learning approaching directions and operation parameters. 
Ni \emph{et al.}~\cite{pointnetppgrasping} directly regress 6D poses after extracting the features of objects using Pointnet++~\cite{pointnetpp}.
6-DoF grasping methods eliminate the restriction on the degree of freedom and generate grasp poses with higher quality. 
However, most of these methods only make use of the partial view point cloud and ignore the color information, while the quality of point cloud is unstable. For dark, transparent and reflective materials, the obtained depth image can be very inaccurate and may lead to failure.  
 
We solve this problem by incorporating both RGB and depth information. With the aid of RGB images, we can alleviate the demand for high quality depth images. Experiments in Section.~\ref{sec:experiments} demonstrate that our method is robust to depth noise and domain variation.

\section{PROBLEM STATEMENT}\label{sec:problemstatement}
\subsection{Definitions}\label{subsec:definitions}
\begin{figure}[htbp]
	\centering
	\includegraphics[width=0.24\textwidth]{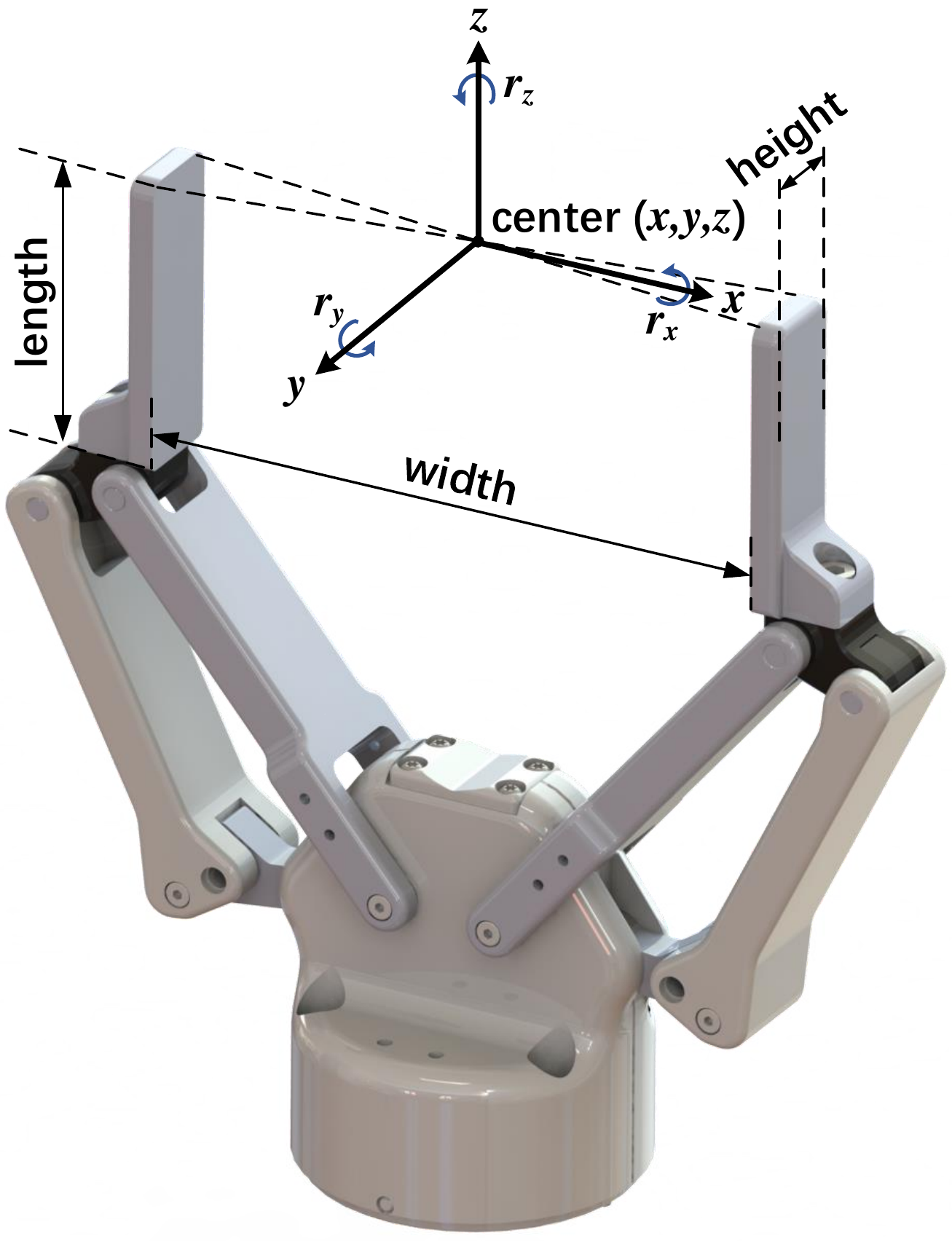}
	\caption{The schematic diagram of a gripper. The length and height of the gripper are fixed according to the gripper configuration. Three translation coordinates, three rotation angles and the width are 7 variables to represent the pose of the gripper.}
	\label{fig:gripper}
\end{figure}

\paragraph{Grasp Pose}
Grasp pose $\mathcal{P}$ is defined by a tuple:
\begin{equation}
    \mathcal{P} = \left(x, y, z, r_x, r_y, r_z, w \right)
\end{equation}
in which $x$, $y$ and $z$ denote the translation of the gripper while $r_x$, $r_y$ and $r_z$ denote the rotation and $w$ denotes the width accordingly.
\paragraph{RGBD Image}
RGBD Image $\mathcal{I}$ is represented by a tuple:
\begin{equation}
    \mathcal{I}  = \left( \mathcal{C}, \mathcal{D}\right)
\end{equation}
in which $\mathcal{C} = \mathbb{R}^{3 \times H \times W } $ denotes the RGB image and $\mathcal{D} = \mathbb{R} ^{H \times W}$ denotes the depth map.
\paragraph{Gripper Configuration}
In this paper, we only take the most common parallel-jaw gripper into consideration. The gripper configuration $\mathcal{G}$ is define by a tuple:
\begin{equation}
    \mathcal{G} = \left( h, l, w_{max}\right)
\end{equation}
in which $h$, $l$ and $w_{max}$ are the height, length and maximum width of the gripper respectively as shown in Figure~\ref{fig:gripper}.
\subsection{Problem Statement}
We make a formal statement of our problem based on the definitions given in Section~\ref{subsec:definitions}. Let $\mathcal{E}$ denote the environment including robot and objects, and $s(\mathcal{E}, \mathcal{I}, \mathcal{G}, \mathcal{P})$ denote a binary variable indicating grasp success or not. A grasp is successful if the object is lifted successfully. Given an RGBD image $\mathcal{I}$ and a gripper configuration $\mathcal{G}$, our goal is to find a set of grasp poses $\mathbf{P}=\left\{\mathcal{P}_1,\mathcal{P}_2,\cdots, \mathcal{P}_k\right\}$ that maximizes the grasp success rate given a fixed $k$:
\begin{equation}
    \left\{\mathcal{P}^{*}_1,\mathcal{P}^{*}_2,\cdots, \mathcal{P}^{*}_k\right\} = \argmax_{|\mathbf{P}|=k} \sum_{\mathcal{P}_i \in \mathbf{P}} \text{Prob}(s=1| \mathcal{E}, \mathcal{I}, \mathcal{G}, \mathcal{P}_i ).
\end{equation}
This means that we hope our algorithm to predict abundant grasp poses to cover the whole scene, so that we can have different candidates for grasp execution.

\section{METHOD}\label{sec:method}
\subsection{Overview}\label{subsec:methodoverview}
As illustrated before, we decouple our problem into two sub-problems. Firstly, \emph{Angle-View Net}(\textbf{AVN}) generates the gripper orientations in different positions of the image:
\begin{equation}
    \mathcal{P}_{img} = \left(u, v, r_x, r_y, r_z, c\right), 
\end{equation}
where $(u,v)$ is a location in image coordinate and $c$ is the success confidence of each prediction.
Secondly, we collect those predictions with high confidence scores and calculate their widths and distances to the image plane, given camera intrinsic parameters and a depth image based on \emph{Fast Analytic Searching}(\textbf{FAS}):
\begin{equation}
    \mathcal{P}_{cam} = \left(x, y, z, r_x, r_y, r_z, w \right)
\end{equation}

\subsection{Angle-View Net}\label{subsec:avn}
The \textbf{AVN} predicts the pixel-wise gripper rotation configurations. One naive way to learn the orientations is to directly regress the rotation matrices or quaternions. However, one location may have several feasible rotations to achieve robust grasping, which makes the regression mechanism impractical.
\begin{figure}[thbp]
	\centering
	\includegraphics[width=0.48\textwidth]{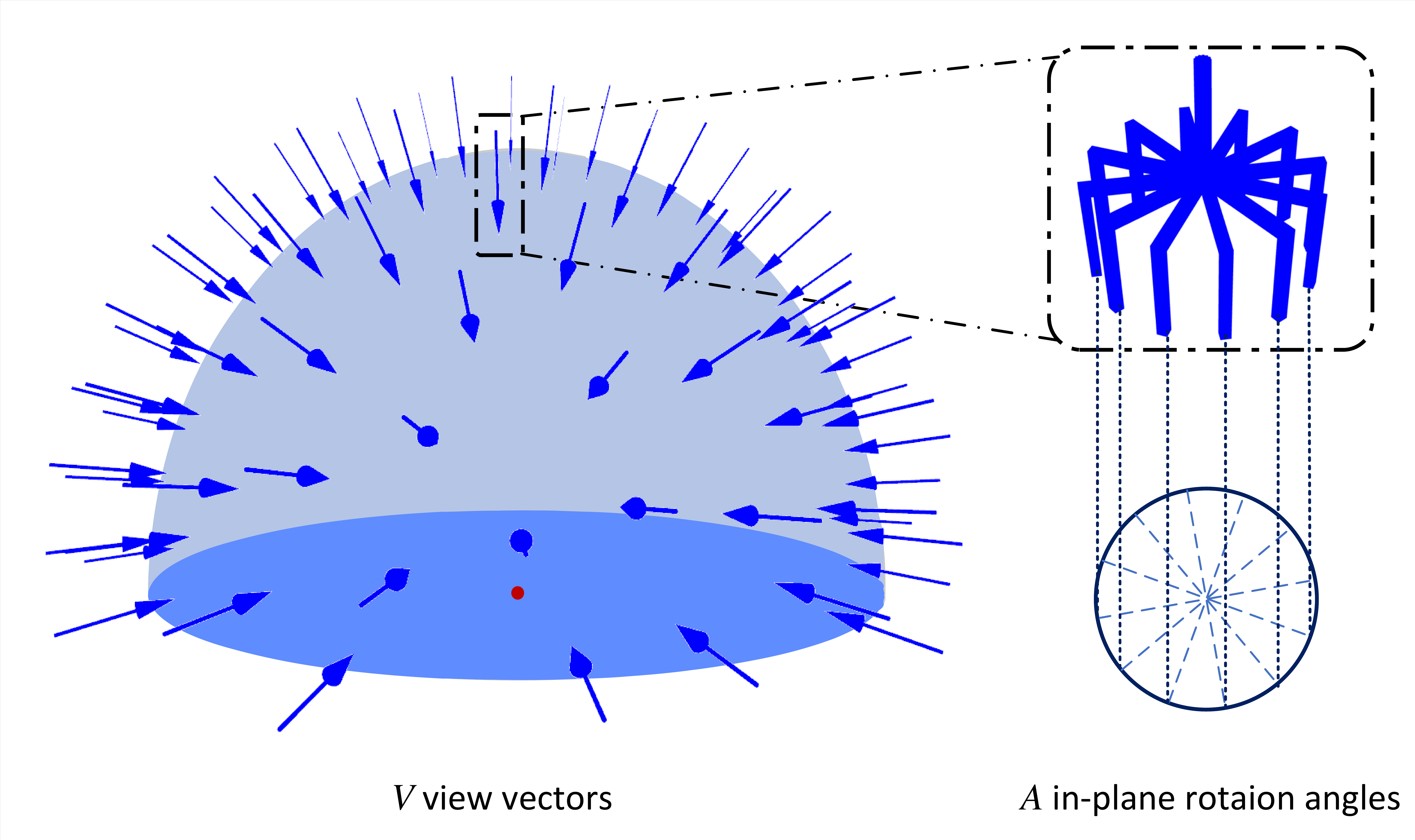}
	\caption{The orientation of the gripper is defined by a view vector and an in-plane rotation angle. As shown in the left part of the figure, $V$ view vectors are uniformly sampled in the upper hemisphere. And as shown in the right part of the figure, $A$ in-plane rotation angles are sampled. In this figure, $V$ and $A$ are set to 120 and 6 respectively.}
	\label{fig:orientation}
\end{figure}

Instead, inspired by GraspNet-1Billion~\cite{graspnet}, we decouple the orientation into the approaching direction and in-plane rotation and treat it as a multi-class classification problem. As shown in Figure~\ref{fig:orientation}, we uniformly sample $V$ views from the upper hemisphere and $A$ in-plane rotation angles. As a result, there are a total of $V \times A$ classes of orientations.

\begin{figure}[t]
	\centering
	\includegraphics[width=0.48\textwidth]{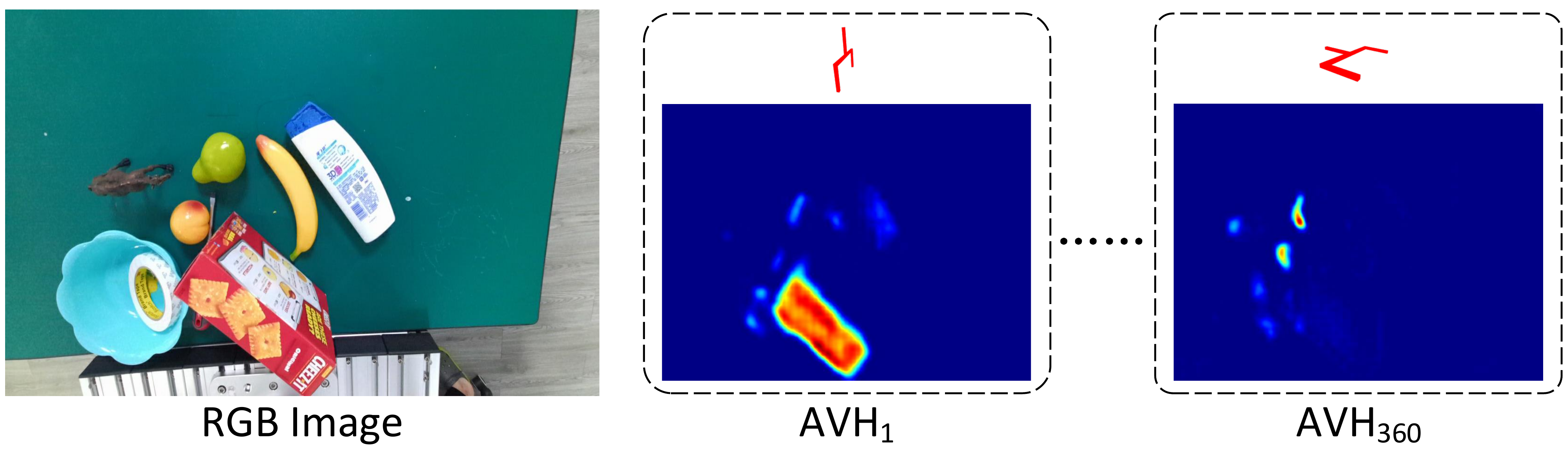}
	\caption{The illustration of two examples of \textbf{AVH}. The left is the original RGB image. The figures on the right are the example \textbf{AVH}s. The corresponding orientation of the gripper is shown above each \textbf{AVH}.}
	\label{fig:avh}
\end{figure}

To obtain the gripper orientations at different locations, we grid the image into $G_H \times G_W$ grids with $G_H$ grids along the vertical direction and $G_W$ grids along the horizontal direction. For each grid, the \textbf{AVN} predicts a 1-dim vector with $V \times A$ elements which indicate the confidence scores of each orientation class in that grid. We denote the final output of \textbf{AVN} as the \emph{Angle-View Heatmap}(\textbf{AVH}) which is defined by a 3D tensor:
\begin{equation}
    \textbf{AVH} = \mathbb{R} ^ {(V * A) \times G_H \times G_W}.
\end{equation}
An example of \textbf{AVH} is given in Figure~\ref{fig:avh}.
\begin{figure}[t]
	\centering
	\includegraphics[width=0.41\textwidth]{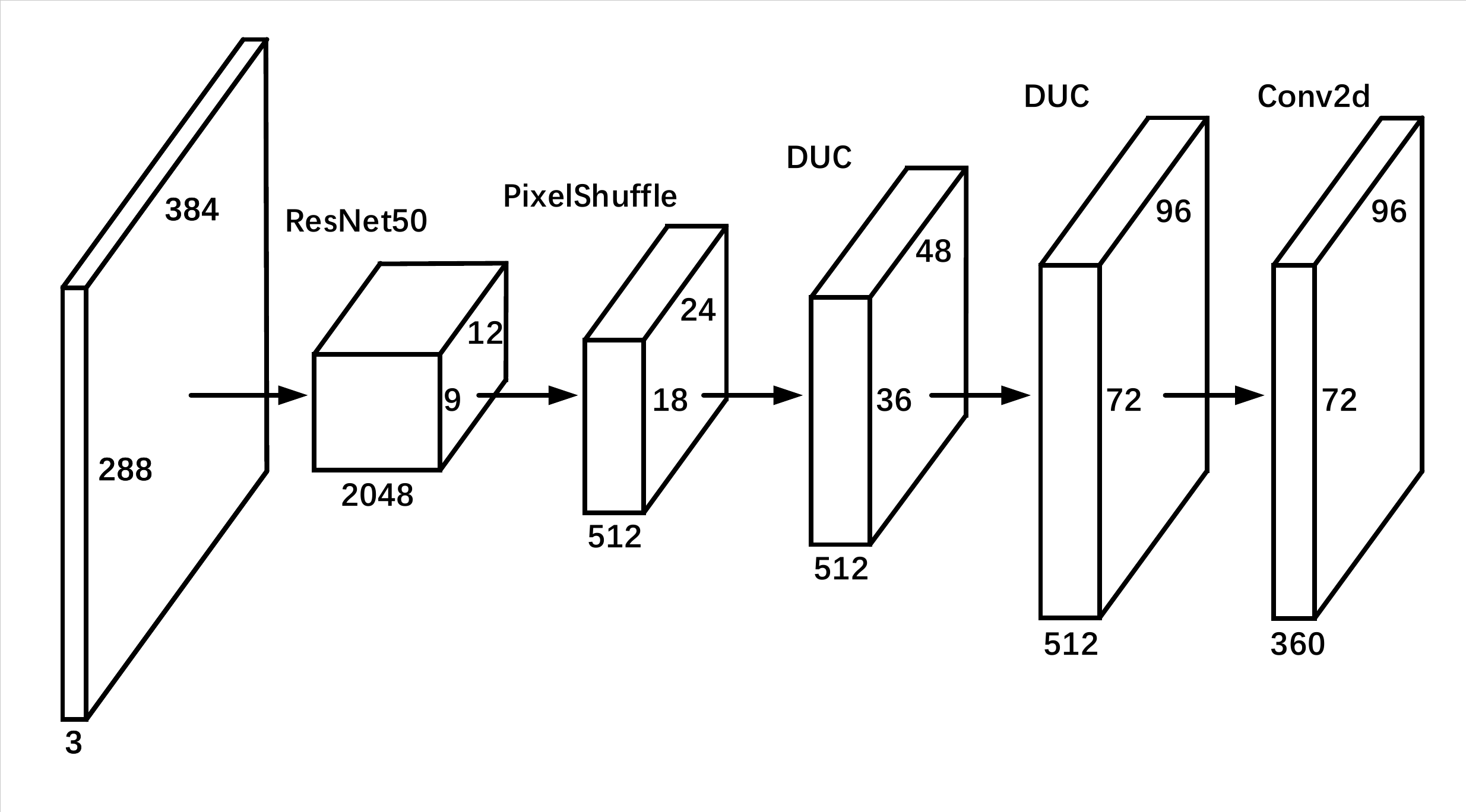}
	\caption{The structure of the Angle-View Net(\textbf{AVN}). The network takes an RGB image as input and passes the image to Resnet to extract dense features. After that, pixel shuffle and Dense Upsampling Convolution(DUC) layers are used to upsampling the features. Finally it generates the \textbf{AVH}.}
	\label{fig:avn}
\end{figure}
To learn the mapping from $\mathbf{C}$ to \textbf{AVH}: $\mathbb{R} ^ {3 \times H \times W} \rightarrow \mathbb{R} ^ {(V * A) \times G_H \times G_W}$, we refer to AlphaPose~\cite{fang2017rmpe, li2019crowdpose} and adopt an encoder-decoder like structure. As illustrated in Figure~\ref{fig:avn}, a ResNet50~\cite{resnet} first encodes the image to high dimensional features. Then a pixel-shuffle layer~\cite{shi2016real} and two dense upsampling convolution(DUC) layers~\cite{duc} decode the features to the \textbf{AVH}. Let $\textbf{AVN}_{\theta}$ denotes $\textbf{AVN}$ with the weights of $\theta$. Given the dataset $\left\{ (\mathbf{C}_1, \textbf{AVH}_1), \cdots, (\mathbf{C}_n, \textbf{AVH}_n) \right\}$ with $(\mathbf{C}_i, \textbf{AVH}_i)$ the $i^{th}$ pair of the RGB image and ground truth \textbf{AVH}, the network can be trained by minimizing the L2 loss function $\mathcal{L}$:
\begin{equation}\label{eqn:loss}
    \theta = \mathop{\arg\min}_{\theta} \sum_{i}^{}{\mathcal{L}(\textbf{AVN}_{\theta}\left(\mathbf{C}_i), \textbf{AVH}_i\right)}, i=(1,2,\cdots,n)
\end{equation}
The acquisition of ground truth \textbf{AVH} for an image will be detailed in Section.~\ref{sec:implementation}.
Compared with those who randomly sample poses, the quality of the generated candidates from $\textbf{AVN}$ are higher and the computing cost for filtering is lower. 
\subsection{Fast Analytic Searching}\label{subsec:fs}
$\textbf{AVN}$ identifies 5 of the 7 DoF of the grasp pose. The width of the gripper and the distance of the gripper to the camera plane remain to be solved. We propose collision and empty grasp detection based Fast Analytic Searching to calculate the width and distance.

One naive way to obtain the width and the distance is learning them by regression or classification. However, learning faces two problems. First, both the width and the distance are dimensional quantities while the RGB values of the image are dimensionless. Thus it's hard to learn the width and the distance without any information of the scale of the image. For example, two images taken at different scales may have the same RGB values. Second, width and distance estimating is closely related to the camera intrinsic parameters. Trained network is hard to generalize to images captured by camera with different intrinsic parameters. 
\begin{figure}[htbp]
    \centering
    \subfigure[Sampling Candidates]
    {
        \begin{minipage}[t]{0.45\textwidth}
        \label{subfig:fs1}
        \centering
        \includegraphics[width=0.70\textwidth]{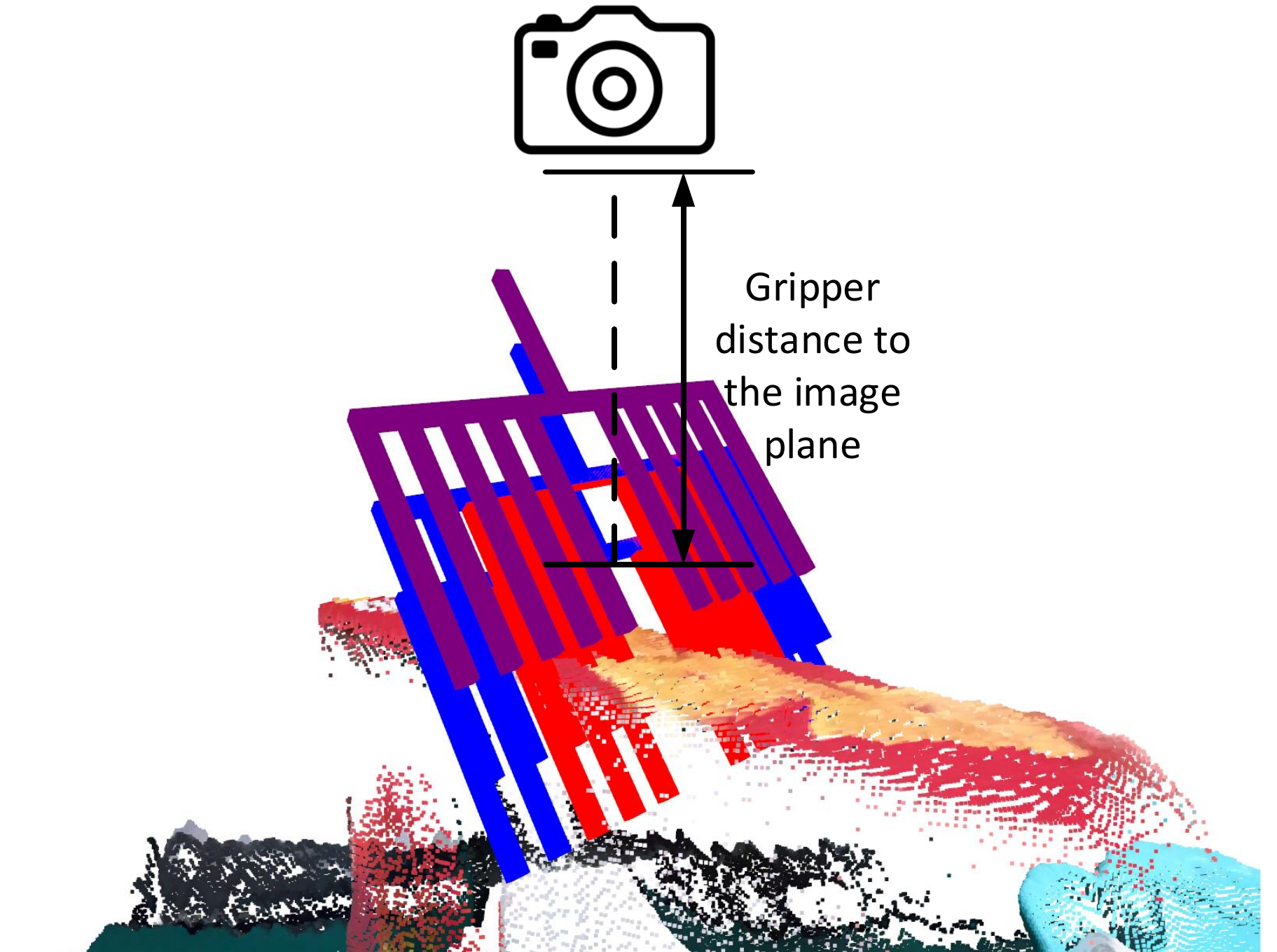}
        \hspace{1mm}
        \end{minipage}
    }
    \subfigure[Result Grasp Pose]
    {
        \begin{minipage}[t]{0.45\textwidth}
        \label{subfig:fs2}
        \centering
        \includegraphics[width=0.70\textwidth]{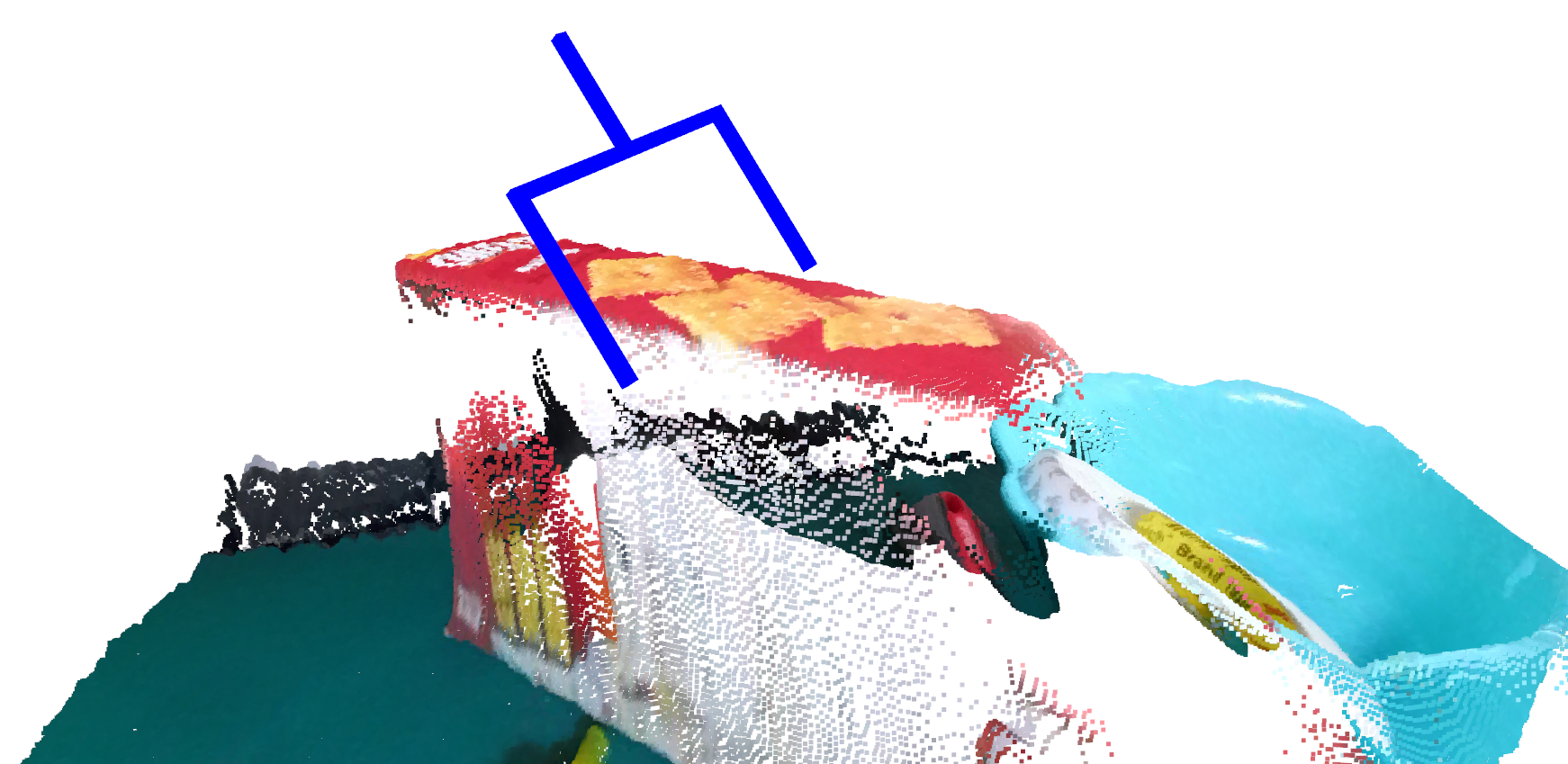}
        \hspace{1mm}
        \end{minipage}
    }
    \caption{The illustration of the \textbf{FAS} Module. Top Figure: The point cloud is reconstructed by the depth image. Candidates with different widths and different distances to the image plane are sampled. We conduct collision and empty checking on these candidates. Red colored candidates collide with the scene point cloud, which violates the first rule. Purple colored candidates have no point in their grasping spaces, which violates the second rule. Blue colored candidates are the remaining ones and considered good grasp poses. Bottom Figure: Among all the good grasp poses, we conduct grasp pose non-maximum suppression to find the one with the largest distance and smallest width.}
    \label{fig:fs}
\end{figure}

Thus instead, we solve the problem by filtering unreasonable grasp poses. As shown in Figure~\ref{fig:fs}, for width estimation, given a gripper configuration $\mathcal{G}$, we uniformly sample $W$ widths from 0 to $w_{max}$. For distance, we assume that the gripper should be close to the corresponding point on partial-view point cloud reconstructed by the depth image. So we sample $D$ distances from the location below the point to location above the point. For each sampled width and distance together with $\mathcal{G}$ and \textbf{AVH}, the gripper can be accurately modeled. As shown in Figure~\ref{fig:fs}, two kinds of grasp poses are filtered out by the rules below.
\begin{enumerate}
    \item There are points in the gripper occupied space.
    \item There is no point in the grasping space. 
\end{enumerate}
\section{IMPLEMENTATION DETAILS}\label{sec:implementation}

\subsection{Dataset}\label{subsec:implementation_dataset}
As mentioned above, we need to generate RGB image and corresponding ground truth \textbf{AVH} pairs for training \textbf{AVN}.
As discussed in Section~\ref{sec:introduction}, Jiang \emph{et al.}~\cite{cornell} and Jacquard~\cite{jacquard} provide the 2D planar grasping datasets. However, all the grasps in these datasets approach the objects in the top-down direction. They are unable to define grasps in all directions. Besides, annotations in these datasets are sparse, which lead to bad performance in generating the ground truth \textbf{AVH}.

We adopt the GraspNet-1Billion~\cite{graspnet} dataset. To the best of our knowledge, this is the only publicly available 6-DoF grasping dataset. It is also easy to generate the ground truth \textbf{AVH} taking the advantage of the dense annotations.

The ground truth \textbf{AVH}s are generated by two steps. Firstly, we initialize an empty $\textbf{AVH} = \mathbf{0} ^{(V * A) \times G_w \times G_H }$ that all the possible angles, views and places are labeled negative. Secondly, for each ground truth grasp pose in the GraspNet-1Billion dataset, we calculate the closest angle, view and place and then label it as positive. For all the 100 training scenes in GraspNet-1Billion dataset, there are about 110 million positive samples and 64 billion negative samples for Kinect and 143 million positive sample and 64 billion negative samples for Intel RealSense captured images.

\subsection{Network and Training}\label{subsec:implementation_train}
As shown in Figure~\ref{fig:avn}, for our network, the input RGB image size is $384 \times 288$ and the output \textbf{AVH} size is $96 \times 72$. We uniformly sample $V=60$ views in the upper hemisphere and $A=6$ angles, generating 360 heatmaps in total. In \textbf{FAS}, the widths are sampled from 1cm to 10cm with the step of 1cm. Depths from the place 2cm below the origin point to the place 2cm above the point with the step of 1cm are explored. 

We train the network using two NVIDIA 2080Ti GPUs with the batchsize of 16 and learning rate of $10^{-3}$ with the ADAM optimizer. The learning rate is decayed by 10 after each 20000 batches. The network is initialized with ImageNet pretrained weights. We train on both Kinect and Intel RealSense camera captured images for 60000 batches. During training, we conduct heavy data augmentation including random crop and color jittering to avoid over-fitting.  

\section{Experiments}\label{sec:experiments}
In this section, we show the experiments carried on both GraspNet-1Billion dataset and a UR-5 robot. We compare our results with several baselines and analyze the differences. 
\vspace{-0.25in}
\subsection{Evaluation Metric}\label{subsec:experiments_metric}
First, we conduct experiments on GraspNet-1Billion dataset. We evaluated our result on the dataset using the standard metric~\cite{graspnet}. This metric first conducts grasp pose non-maximum suppression(GPNMS)~\cite{graspnet_web} to avoid duplicated grasp poses in a small region.
Then, for the top $k$ candidates, it labels those collided grasp poses as negative predictions based on the reconstructed scene.
A binary label is assigned to the remaining grasp poses given different friction coefficients by force-closure metric~\cite{dexnet2}.
Finally, it calculates the \textbf{AP} by finding the mean average precision of different friction coefficients and different $k$ range from 1 to 50. 
\vspace{-0.05in}
\subsection{Results on GraspNet-1Billion}\label{subsec:experiments_result_graspnet}
\subsubsection{Baselines} We adopt GG-CNN~\cite{ggcnn}, Chu \emph{et al.}~\cite{multi}, Fang \emph{et al.}~\cite{graspnet} as baselines. We compare both the quantitative results and visualization results to show the effectiveness of our RGBD-Grasp.
For GG-CNN~\cite{ggcnn} and Chu \emph{et al.}~\cite{multi}, we first convert the GraspNet-1Billion data into the Cornell dataset~\cite{cornell} format by selecting top-down grasps and projecting them on the image. Then, for Chu \emph{et al.}~\cite{multi}, we retrain the network using the converted GraspNet-1Billion dataset. For GG-CNN~\cite{ggcnn}, the retrained model gets poor performance and we use the pretrained model on Cornell dataset for evaluation instead. To show the effectiveness of our \textbf{AVN}, we also conduct an ablation study by removing it from our pipeline and randomly sample candidates for \textbf{FAS}.

To verify the robustness of our method towards depth images with poor quality, we randomly add some Gaussian noise on the depth images and test the performances of RGBD-Grasp and Fang \emph{et al.}~\cite{graspnet}.
Besides, we have also conducted cross-domain testing by training the network on one of the two cameras between Kinect and RealSense and evaluating the result on the other one. This is to verify the cross domain generalization ability of algorithms.
\begin{figure}[t]
	\centering
	\includegraphics[width=0.40\textwidth]{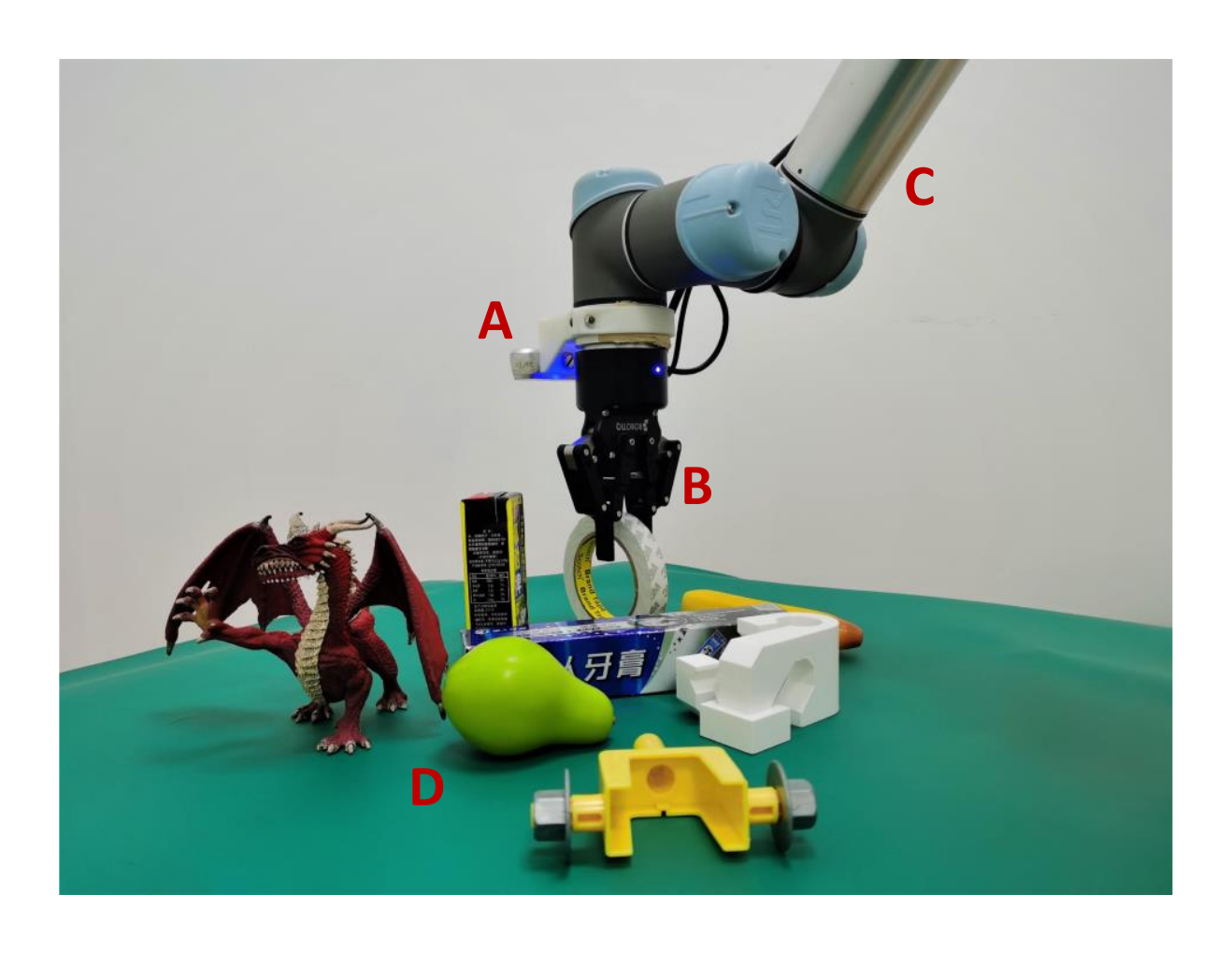}
	\caption{Real robot experiments on a cluttered scene. A: Intel RealSense RGBD camera. B: Robotiq two-finger gripper. C: UR-5 Robot. D: Objects from GraspNet 1-Billion dataset. The robot executes the best grasp pose output by RGBD-Grasp and the objects are grabbed one by one.} 
	\label{fig:robot}
\end{figure}
\begin{table*}[]
    \centering
    \begin{tabular}{|c|ccc|ccc|ccc|}
        \hline
        \multirow{2}{*}{Methods} & \multicolumn{3}{c|}{Seen} & \multicolumn{3}{c|}{Unseen} & \multicolumn{3}{c|}{Novel} \\
        \cline{2-10}
        & \textbf{AP} & $\text{AP}_{\textbf{0.8}}$ & $\text{AP}_{\textbf{0.4}}$ & \textbf{AP} & $\text{AP}_{\textbf{0.8}}$ & $\text{AP}_{\textbf{0.4}}$ & \textbf{AP} & $\text{AP}_{\textbf{0.8}}$ & $\text{AP}_{\textbf{0.4}}$ \\
        \hline
        \footnotesize{Random } & \footnotesize{10.87/11.44} & \footnotesize{12.97/13.43} & \footnotesize{9.66/10.01} & \footnotesize{10.45/11.10} & \footnotesize{12.22/12.98} & \footnotesize{8.94/9.76} & \footnotesize{4.90/5.54} & \footnotesize{5.76/6.76} & \footnotesize{2.87/3.16} \\
        \footnotesize{GG-CNN~\cite{ggcnn}} & \footnotesize{15.48/16.89}  & \footnotesize{21.84/22.47} & \footnotesize{10.25/11.23} & \footnotesize{13.26/15.05} & \footnotesize{18.37/19.76} & \footnotesize{4.62/6.19} & \footnotesize{5.52/7.38} & \footnotesize{5.93/8.78} & \footnotesize{1.86/1.32} \\
        \footnotesize{Chu \emph{et al.}~\cite{multi}} & \footnotesize{15.97/17.59} & \footnotesize{23.66/24.67} & \footnotesize{10.80/12.74} & \footnotesize{15.41/17.36} & \footnotesize{20.21/21.64} & \footnotesize{7.06/8.86} & \footnotesize{7.64/8.04} & \footnotesize{8.69/9.34} & \footnotesize{2.52/1.76} \\

        \footnotesize{Fang \emph{et al.}~\cite{graspnet}} & \footnotesize{27.56/29.88} & \footnotesize{33.43/36.19} & \footnotesize{16.95/19.31} & \footnotesize{26.11/27.84} & \footnotesize{34.18/33.19} & \footnotesize{14.23/16.62} & \footnotesize{10.55/11.51}  & \footnotesize{11.25/12.92} & \footnotesize{3.98/3.56}\\
        \hline
        \footnotesize{~\cite{graspnet} with noise} & 
        \footnotesize{22.35/24.35} & \footnotesize{27.96/30.34} & \footnotesize{12.23/14.24} & \footnotesize{21.12/21.34} & \footnotesize{28.38/27.90} & \footnotesize{11.23/12.30} & \footnotesize{7.23/8.10}  & \footnotesize{8.02/9.87} & \footnotesize{2.52/2.23}\\
        \footnotesize{Ours with noise} & 
        \footnotesize{26.34/28.98} & \footnotesize{32.78/35.91} & \footnotesize{15.84/18.40} & \footnotesize{27.01/29.01} & \footnotesize{35.35/34.21} & \footnotesize{15.10/17.32} & \footnotesize{10.95/11.64}  & \footnotesize{11.15/12.62} & \footnotesize{4.53/4.68}\\
        \hline
        \footnotesize{Ours} & 
        \footnotesize{\textbf{27.98}/\textbf{32.08}} & \footnotesize{\textbf{33.47}/\textbf{39.46}} & \footnotesize{\textbf{17.75}/\textbf{20.85}} & \footnotesize{\textbf{27.23}/\textbf{30.40}} & \footnotesize{\textbf{36.34}/\textbf{37.87}} & \footnotesize{\textbf{15.60}/\textbf{18.72}} & \footnotesize{\textbf{12.25}/\textbf{13.08}}  & \footnotesize{\textbf{12.45}/\textbf{13.79}} & \footnotesize{\textbf{5.62}/\textbf{6.01}}\\
        \hline
    \end{tabular}
    \vspace{1mm}
    \caption{Evaluation result for different methods. The table shows the results on data captured by RealSense/Kinect respectively.}
    \label{tab:result}
    \vspace{-0.2in}
\end{table*}
\subsubsection{Results and Analysis} 
\begin{figure*}[t]
	\centering
	\includegraphics[width=0.984\textwidth]{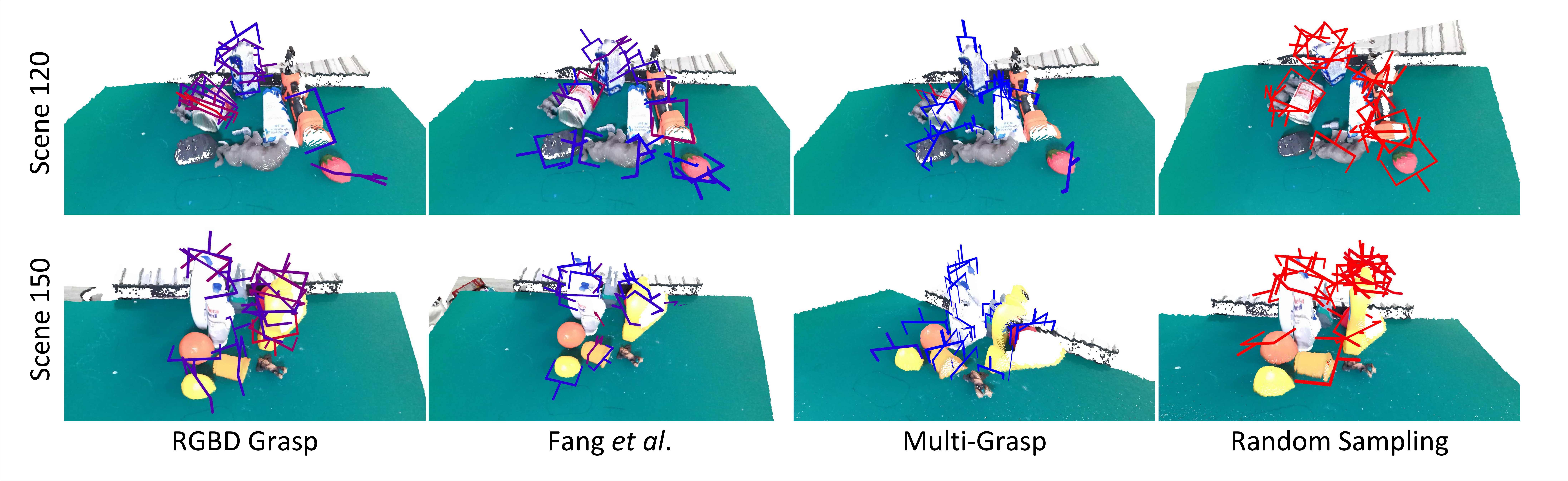}
	\caption{Example visualization results of four different algorithms on scene 120 and scene 150 in GraspNet-1Billion dataset for Kinect captured image. The color of the gripper tells the confidence score for each grasp. For more confident grasp pose, the gripper color is closer to red. }
	\label{fig:result}
\end{figure*}

The scores of several methods on the GraspNet-1Billion metric~\cite{graspnet} are shown in Table~\ref{tab:result} and some of the examples are visualized in Figure~\ref{fig:result}. The definitions of \textbf{APs} can be found in \cite{graspnet}.

As for 2D planar grasping methods~\cite{ggcnn}~\cite{multi}, the restriction on DoF greatly harms their performances. And for random sampling, it outputs few reasonable grasp poses as the orientations of the grasps are randomly selected even though the \textbf{FAS} module filters some of the bad grasp poses.

RGBD-Grasp slightly outperforms Fang \emph{et al.} on the original dataset. Moreover, when it comes to feeding depth images with noise, the score obtained by Fang \emph{et al.} drops significantly while RGBD-Grasp only gets a small decrease. 

As shown in Table~\ref{tab:cross_domain}. RGBD-Grasp performs well even though it's trained on a different camera. However, Fang \emph{et al.} relies heavily on the depth image and as a result gets low scores when evaluated in a cross-camera setting. 

\begin{table}[]
    \centering
    \begin{tabular}{|c|c|c|ccc|}
        \hline
        \textbf{Methods} & \textbf{Train on} & \textbf{Eval on} & \textbf{Seen} & \textbf{Similar} & \textbf{Novel} \\
        \hline
        \footnotesize{Fang \emph{et al.}} & \footnotesize{Kinect} & \footnotesize{Kinect}& \footnotesize{29.88} & \footnotesize{27.84} & \footnotesize{11.51} \\
        \hline
        \footnotesize{Fang \emph{et al.}} & \footnotesize{RealSense} & \footnotesize{Kinect}& \footnotesize{22.00} & \footnotesize{21.52} & \footnotesize{8.40} \\
        \hline
        \footnotesize{Fang \emph{et al.}} & \footnotesize{Realsense} & \footnotesize{RealSense}& \footnotesize{27.56} & \footnotesize{26.11} & \footnotesize{10.55} \\
        \hline

        \footnotesize{Fang \emph{et al.}} & \footnotesize{Kinect} & \footnotesize{RealSense}& \footnotesize{20.19} & \footnotesize{18.97} & \footnotesize{7.97} \\
        \hline
        \footnotesize{RGBD Grasp} & \footnotesize{Kinect} & \footnotesize{Kinect}& \footnotesize{32.08} & \footnotesize{30.40} & \footnotesize{13.08} \\
        \hline
        \footnotesize{RGBD Grasp} & \footnotesize{RealSense} & \footnotesize{Kinect}& \footnotesize{27.25} & \footnotesize{26.54} & \footnotesize{11.25} \\
        \hline
        \footnotesize{RGBD Grasp} & \footnotesize{Realsense} & \footnotesize{RealSense}& \footnotesize{27.98} & \footnotesize{27.23} & \footnotesize{12.25} \\
        \hline

        \footnotesize{RGBD Grasp} & \footnotesize{Kinect} & \footnotesize{RealSense}& \footnotesize{26.65} & \footnotesize{27.81} & \footnotesize{13.15} \\   
        \hline
    \end{tabular}
    \vspace{1mm}
    \caption{Cross Domain Testing Results.}
    \label{tab:cross_domain}
    \vspace{-0.2in}
\end{table}
\vspace{-0.05in}
\subsection{Real Robot Experiments}\label{subsec:experiments_real}
We also conduct real robot experiments. As shown in Figure~\ref{fig:robot}, the experiment is carried on a UR-5 robot mounted with an Intel RealSense camera and a Robotiq two-finger gripper. A computer with an NVIDIA GTX 2060 GPU, an Intel i7-8750H CPU, 8G RAM and an Ubuntu 18.04 operating system is used to run our pipeline and control the robot. The trajectory is planed using MoveIt~\cite{moveit} to avoid collision among the robot, gripper, camera and objects.

\subsubsection{Experiment on Single Object Scenes}

We randomly select 9 objects from GraspNet-1Billion dataset and place them on the table at random place with random orientation. Objects with different difficulties are selected to analyze the performance of generalization. We conduct 20 experiments for each object and calculate the success rate. The results are shown in Table~\ref{tab:single_result}.
\begin{table}[]
\centering
\begin{tabular}{|c|c|c|c|c|c|}
\hline
ID &Name & Type & Attempt & Success &  Success Rate  \\
\hline
1 & 004\_sugar\_box & seen & 20 & 19 & 95\% \\
\hline
5 & 011\_banana & seen & 20 & 20 & 100\%  \\
\hline
12 & 013\_apple & seen & 20 & 19 & 95\%  \\
\hline
38 & dabao\_sod & similar & 20 & 19 & 95\%  \\
\hline
39 & soap\_box & similar & 20 & 19 & 95\% \\
\hline
43 & baoke\_marker & similar & 20 & 18 & 90\%  \\
\hline
72 & peeler\_cover & novel & 20 & 17 & 85\%  \\
\hline
33 & dragon & novel & 20 & 18 & 90\%  \\
\hline
74 & ice\_cube\_mould & novel & 20 & 16 & 80\% \\
\hline

- & Average & - & 20 & 18.33 & 91.67\%\\
\hline

\end{tabular}
\caption{Results of Real Robot Experiments on Single Object Scenes}
\label{tab:single_result}
\vspace{-0.2in}
\end{table}
\subsubsection{Experiment on Cluttered Scenes}
We also conduct experiments on cluttered scenes. In each scene, we randomly select 4-8 objects from GraspNet-1Billion and randomly place them on the table. We find and execute the best grasp pose candidate using the pipeline until all the objects are taken away. We define the success rate of cluttered-scene experiments by:

$$success\ rate = \dfrac{number\ of\ objects}{number\ of\ attempts}$$
\begin{table}[]
\centering
\begin{tabular}{|c|c|c|c|}
\hline

Objects ID & Number & Attempts & Success Rate  \\
\hline
5, 15, 31, 33, 58, 68, 70, 75 &8 & 8 & 100.0\%  \\
\hline
1, 2, 4, 5, 13 & 5 & 6 & 83.3\%  \\
\hline
17, 21, 26, 24, 32, 35, 36, 37 & 8 & 10 & 80.0\%  \\
\hline
35, 36, 47, 50, 55, 59 & 6 & 6 & 100.0\%  \\
\hline
5, 15, 26, 50, 72, 73, 77 & 7 & 8 & 87.5\%  \\
\hline
6, 13, 19, 21, 68, 70, 74 &  7 & 7 & 100.0\%  \\
\hline
Average & 6.83 & 7.5 & 91.1\%  \\
\hline

\end{tabular}
\caption{Results of Real Robot Experiments on Cluttered Scenes}
\label{tab:scene_result}
\vspace{-0.2in}
\end{table}
The experiment results are shown in Table~\ref{tab:scene_result}. We can see that RGBD-Grasp can achieve satisfactory success rate.
\section{Conclusion}
In this paper, we propose a novel 7-DoF grasp pose detection pipeline named RGBD-Grasp. It decouples the problem of finding grasp poses into two sub-questions. The \emph{Angle-View Net}(\textbf{AVN}) module takes a monocular RGB image as input and generates heatmaps to predict both grasping locations in the image plane and the orientations of the gripper. The \emph{Fast Analytic Searching}(\textbf{FAS}) module takes the heatmaps and a depth image as input and outputs the widths and distances from the gripper centers to the image plane. We achieve state-of-the-art results on GraspNet-1Billion benchmark. We also conduct experiments using a UR-5 robot with a Robotiq two-finger gripper on both single object scenes and cluttered scenes. Both experiments report high success rates and therefore verify the effectiveness of RGBD-Grasp. 
\section{ACKNOWLEDGMENT}
This work is supported in part by the National Key R\&D Program of China, No. 2017YFA0700800, National Natural Science Foundation of China under Grants 61772332, Shanghai Qi Zhi Institute, SHEITC(2018-RGZN-02046) and Baidu Fellowship.

\printbibliography

@inproceedings{shi2016real,
  title={Real-time single image and video super-resolution using an efficient sub-pixel convolutional neural network},
  author={Shi, Wenzhe and Caballero, Jose and Husz{\'a}r, Ferenc and Totz, Johannes and Aitken, Andrew P and Bishop, Rob and Rueckert, Daniel and Wang, Zehan},
  booktitle={Proceedings of the IEEE conference on computer vision and pattern recognition},
  year={2016}
}

@inproceedings{fang2017rmpe,
  title={Rmpe: Regional multi-person pose estimation},
  author={Fang, Hao-Shu and Xie, Shuqin and Tai, Yu-Wing and Lu, Cewu},
  booktitle={Proceedings of the IEEE International Conference on Computer Vision},
  year={2017}
}

@article{2014redmon,
  title={Real-Time Grasp Detection Using Convolutional Neural Networks},
  author={ Redmon, Joseph  and  Angelova, Anelia },
  journal={Proceedings IEEE International Conference on Robotics and Automation (ICRA)},
  year={2014},
}

@inproceedings{dsgd,
  title={Densely supervised grasp detector (DSGD)},
  author={Asif, Umar and Tang, Jianbin and Harrer, Stefan},
  booktitle={Proceedings of the AAAI Conference on Artificial Intelligence},
  volume={33},
  pages={8085--8093},
  year={2019}
}

@inproceedings{graspnet,
  title={GraspNet-1Billion: A Large-Scale Benchmark for General Object Grasping},
  author={Fang, Hao-Shu and Wang, Chenxi and Gou, Minghao and Lu, Cewu},
  booktitle={Proceedings of the IEEE/CVF Conference on Computer Vision and Pattern Recognition (CVPR)},
  pages={11444--11453},
  year={2020}
}

@article{review1,
  title={Vision-based robotic grasping from object localization, object pose estimation to grasp estimation for parallel grippers: a review},
  author={Du, Guoguang and Wang, Kai and Lian, Shiguo and Zhao, Kaiyong},
  journal={Artificial Intelligence Review},
  pages={1--58},
  year={2020},
  publisher={Springer}
}

@article{ggcnn,
  title={Closing the loop for robotic grasping: A real-time, generative grasp synthesis approach},
  author={Morrison, Douglas and Corke, Peter and Leitner, Jurgen},
  journal={Robotics: Science and Systems XIV},
  pages={1--10},
  year={2018},
  publisher={Robotics Science and Systems Foundation}
}

@article{zhao2018estimating,
  title={Estimating 6d pose from localizing designated surface keypoints},
  author={Zhao, Zelin and Peng, Gao and Wang, Haoyu and Fang, Hao-Shu and Li, Chengkun and Lu, Cewu},
  journal={arXiv preprint arXiv:1812.01387},
  year={2018}
}

@inproceedings{he2020pvn3d,
  title={Pvn3d: A deep point-wise 3d keypoints voting network for 6dof pose estimation},
  author={He, Yisheng and Sun, Wei and Huang, Haibin and Liu, Jianran and Fan, Haoqiang and Sun, Jian},
  booktitle={Proceedings of the IEEE/CVF conference on computer vision and pattern recognition},
  pages={11632--11641},
  year={2020}
}

@inproceedings{peng2019pvnet,
  title={Pvnet: Pixel-wise voting network for 6dof pose estimation},
  author={Peng, Sida and Liu, Yuan and Huang, Qixing and Zhou, Xiaowei and Bao, Hujun},
  booktitle={Proceedings of the IEEE/CVF Conference on Computer Vision and Pattern Recognition},
  pages={4561--4570},
  year={2019}
}

@ARTICLE{review2,
  author={M. Q. {Mohammed} and K. L. {Chung} and C. S. {Chyi}},
  journal={IEEE Access}, 
  title={Review of Deep Reinforcement Learning-Based Object Grasping: Techniques, Open Challenges, and Recommendations}, 
  year={2020},
  volume={8},
  number={},
  pages={178450-178481},}

@inproceedings{li2019crowdpose,
  title={Crowdpose: Efficient crowded scenes pose estimation and a new benchmark},
  author={Li, Jiefeng and Wang, Can and Zhu, Hao and Mao, Yihuan and Fang, Hao-Shu and Lu, Cewu},
  booktitle={Proceedings of the IEEE/CVF Conference on Computer Vision and Pattern Recognition},
  pages={10863--10872},
  year={2019}
}

@inproceedings{asif,
  title={GraspNet: An Efficient Convolutional Neural Network for Real-time Grasp Detection for Low-powered Devices.},
  author={Asif, Umar and Tang, Jianbin and Harrer, Stefan},
  booktitle={IJCAI},
  pages={4875--4882},
  year={2018}
}

@article{multi,
  title={Real-world multiobject, multigrasp detection},
  author={Chu, Fu-Jen and Xu, Ruinian and Vela, Patricio A},
  journal={IEEE Robotics and Automation Letters (RA-L)},
  volume={3},
  number={4},
  pages={3355--3362},
  year={2018},
  publisher={IEEE}
}

@inproceedings{pointnetpp,
  title={Pointnet++: Deep hierarchical feature learning on point sets in a metric space},
  author={Qi, Charles Ruizhongtai and Yi, Li and Su, Hao and Guibas, Leonidas J},
  booktitle={Advances in neural information processing systems (NeurIPS},
  pages={5099--5108},
  year={2017}
}

@inproceedings{s4g,
  title={S4g: Amodal single-view single-shot se (3) grasp detection in cluttered scenes},
  author={Qin, Yuzhe and Chen, Rui and Zhu, Hao and Song, Meng and Xu, Jing and Su, Hao},
  booktitle={Conference on robot learning (CoRL)},
  pages={53--65},
  year={2020},
  organization={PMLR}
}

@article{review3,
  title={A Survey on Learning-Based Robotic Grasping},
  author={Kleeberger, Kilian and Bormann, Richard and Kraus, Werner and Huber, Marco F},
  journal={Current Robotics Reports},
  pages={1--11},
  year={2020},
  publisher={Springer}
}

@article{review4,
  title={Review of deep learning methods in robotic grasp detection},
  author={Caldera, Shehan and Rassau, Alexander and Chai, Douglas},
  journal={Multimodal Technologies and Interaction},
  volume={2},
  number={3},
  pages={57},
  year={2018},
  publisher={Multidisciplinary Digital Publishing Institute}
}

@inproceedings{dexnet2,
    AUTHOR={Jeffrey Mahler AND Jacky Liang AND Sherdil Niyaz AND Michael Laskey AND Richard Doan AND Xinyu Liu AND Juan Aparicio AND Ken Goldberg}, 
    TITLE={Dex-Net 2.0: Deep Learning to Plan Robust Grasps with Synthetic Point Clouds and Analytic Grasp Metrics}, 
    BOOKTITLE={Proceedings of Robotics: Science and Systems (RSS)}, 
    year={2017}, 
    ADDRESS={Cambridge, Massachusetts}, 
    MONTH={July}, 
    DOI={10.15607/RSS.2017.XIII.058} 
}

@inproceedings{cornell,
  title={Efficient grasping from rgbd images: Learning using a new rectangle representation},
  author={Jiang, Yun and Moseson, Stephen and Saxena, Ashutosh},
  booktitle={2011 IEEE International conference on robotics and automation (ICRA)},
  pages={3304--3311},
  year={2011},
  organization={IEEE}
}

@inproceedings{pointnetgpd,
  title={Pointnetgpd: Detecting grasp configurations from point sets},
  author={Liang, Hongzhuo and Ma, Xiaojian and Li, Shuang and G{\"o}rner, Michael and Tang, Song and Fang, Bin and Sun, Fuchun and Zhang, Jianwei},
  booktitle={2019 International Conference on Robotics and Automation (ICRA)},
  pages={3629--3635},
  year={2019},
  organization={IEEE}
}

@article{gpd,
  title={Grasp pose detection in point clouds},
  author={ten Pas, Andreas and Gualtieri, Marcus and Saenko, Kate and Platt, Robert},
  journal={The International Journal of Robotics Research (IJRR)},
  volume={36},
  number={13-14},
  pages={1455--1473},
  year={2017},
  publisher={SAGE Publications Sage UK: London, England}
}

@inproceedings{supersizing,
  title={Supersizing self-supervision: Learning to grasp from 50k tries and 700 robot hours},
  author={Pinto, Lerrel and Gupta, Abhinav},
  booktitle={2016 IEEE international conference on robotics and automation (ICRA)},
  pages={3406--3413},
  year={2016},
  organization={IEEE}
}

@inproceedings{6dofgraspnet,
  title={6-dof graspnet: Variational grasp generation for object manipulation},
  author={Mousavian, Arsalan and Eppner, Clemens and Fox, Dieter},
  booktitle={Proceedings of the IEEE International Conference on Computer Vision (ICCV)},
  pages={2901--2910},
  year={2019}
}

@inproceedings{kumra,
  title={Robotic grasp detection using deep convolutional neural networks},
  author={Kumra, Sulabh and Kanan, Christopher},
  booktitle={2017 IEEE/RSJ International Conference on Intelligent Robots and Systems (IROS)},
  pages={769--776},
  year={2017},
  organization={IEEE}
}

@article{denseobjectnets,
  title={Dense Object Nets: Learning Dense Visual Object Descriptors By and For Robotic Manipulation},
  author={Florence, Peter and Manuelli, Lucas and Tedrake, Russ},
  journal={Conference on Robot Learning (CoRL)},
  year={2018}
}

@inproceedings{tian2019transferring,
  title={Transferring Grasp Configurations using Active Learning and Local Replanning},
  author={Tian, Hao and Wang, Changbo and Manocha, Dinesh and Zhang, Xinyu},
  booktitle={2019 International Conference on Robotics and Automation (ICRA)},
  pages={1622--1628},
  year={2019},
  organization={IEEE}
}

@article{lenz,
  title={Deep learning for detecting robotic grasps},
  author={Lenz, Ian and Lee, Honglak and Saxena, Ashutosh},
  journal={The International Journal of Robotics Research (IJRR)},
  volume={34},
  number={4-5},
  pages={705--724},
  year={2015},
  publisher={SAGE Publications Sage UK: London, England}
}

@inproceedings{fasterrcnn,
  title={Faster r-cnn: Towards real-time object detection with region proposal networks},
  author={Ren, Shaoqing and He, Kaiming and Girshick, Ross and Sun, Jian},
  booktitle={Advances in neural information processing systems (NeurIPS)},
  pages={91--99},
  year={2015}
}

@ARTICLE{dgcmnet,
AUTHOR={Patten, Timothy and Park, Kiru and Vincze, Markus},   
TITLE={DGCM-Net: Dense Geometrical Correspondence Matching Network for Incremental Experience-Based Robotic Grasping},      
JOURNAL={Frontiers in Robotics and AI},      
VOLUME={7},      
PAGES={120},     
YEAR={2020},      
ISSN={2296-9144},   
}

@inproceedings{shapecompletion,
  title={Shape completion enabled robotic grasping},
  author={Varley, Jacob and DeChant, Chad and Richardson, Adam and Ruales, Joaqu{\'\i}n and Allen, Peter},
  booktitle={2017 IEEE/RSJ international conference on intelligent robots and systems (IROS)},
  pages={2442--2447},
  year={2017},
  organization={IEEE}
}

@INPROCEEDINGS{Lundell,
  author={J. {Lundell} and F. {Verdoja} and V. {Kyrki}},
  booktitle={2019 IEEE/RSJ International Conference on Intelligent Robots and Systems (IROS)}, 
  title={Robust Grasp Planning Over Uncertain Shape Completions}, 
  year={2019},
  volume={},
  number={},
  pages={1526-1532},}

@inproceedings{yan2018learning,
  title={Learning 6-dof grasping interaction via deep geometry-aware 3d representations},
  author={Yan, Xinchen and Hsu, Jasmined and Khansari, Mohammad and Bai, Yunfei and Pathak, Arkanath and Gupta, Abhinav and Davidson, James and Lee, Honglak},
  booktitle={2018 IEEE International Conference on Robotics and Automation (ICRA)},
  pages={1--9},
  year={2018},
  organization={IEEE}
}

@inproceedings{traditional1,
  title={Robotic grasping and contact: A review},
  author={Bicchi, Antonio and Kumar, Vijay},
  booktitle={IEEE International Conference on Robotics and Automation (ICRA)},
  volume={1},
  pages={348--353},
  year={2000},
  organization={IEEE}
}

@inproceedings{traditional2,
  title={Semantic grasping: Planning robotic grasps functionally suitable for an object manipulation task},
  author={Dang, Hao and Allen, Peter K},
  booktitle={2012 IEEE/RSJ International Conference on Intelligent Robots and Systems (IROS)},
  pages={1311--1317},
  year={2012},
  organization={IEEE}
}

@ARTICLE{traditional3,
  author={J. {Bohg} and A. {Morales} and T. {Asfour} and D. {Kragic}},
  journal={IEEE Transactions on Robotics (T-RO)}, 
  title={Data-Driven Grasp Synthesis—A Survey}, 
  year={2014},
  volume={30},
  number={2},
  pages={289-309},
 }

@INPROCEEDINGS{jacquard,
  author={A. {Depierre} and E. {Dellandréa} and L. {Chen}},
  booktitle={2018 IEEE/RSJ International Conference on Intelligent Robots and Systems (IROS)}, 
  title={Jacquard: A Large Scale Dataset for Robotic Grasp Detection}, 
  year={2018},
  volume={},
  number={},
  pages={3511-3516},}

@inproceedings{duc,
  title={Understanding convolution for semantic segmentation},
  author={Wang, Panqu and Chen, Pengfei and Yuan, Ye and Liu, Ding and Huang, Zehua and Hou, Xiaodi and Cottrell, Garrison},
  booktitle={2018 IEEE winter conference on applications of computer vision (WACV)},
  pages={1451--1460},
  year={2018},
  organization={IEEE}
}

@inproceedings{resnet,
  title={Deep residual learning for image recognition},
  author={He, Kaiming and Zhang, Xiangyu and Ren, Shaoqing and Sun, Jian},
  booktitle={Proceedings of the IEEE conference on computer vision and pattern recognition (CVPR)},
  pages={770--778},
  year={2016}
}

@article{moveit,
  title={Reducing the barrier to entry of complex robotic software: a moveit! case study},
  author={Coleman, David and Sucan, Ioan and Chitta, Sachin and Correll, Nikolaus},
  journal={arXiv preprint arXiv:1404.3785},
  year={2014}
}

@INPROCEEDINGS{pointnetppgrasping,
  author={P. {Ni} and W. {Zhang} and X. {Zhu} and Q. {Cao}},
  booktitle={2020 IEEE International Conference on Robotics and Automation (ICRA)}, 
  title={PointNet++ Grasping: Learning An End-to-end Spatial Grasp Generation Algorithm from Sparse Point Clouds}, 
  year={2020},
  volume={},
  number={},
  pages={3619-3625}
}

@misc{graspnet_web,
year = {2020},
note = {\url{https://graspnet.net/evaluation.html}}
}

\end{document}